\documentclass[runningheads]{llncs}

\usepackage{eccv}

\usepackage{eccvabbrv}

\usepackage{graphicx}
\usepackage{booktabs}
\usepackage{enumerate}
\usepackage{marvosym}
\usepackage[accsupp]{axessibility}  % Improves PDF readability for those with disabilities.

\usepackage[breaklinks,colorlinks,citecolor=eccvblue]{hyperref}
\usepackage{orcidlink}

\begin{document}

% ---------------------------------------------------------------
% TODO REVIEW: Replace with your title
\title{PFGS: High Fidelity Point Cloud Rendering via Feature Splatting} 

% TODO REVIEW: If the paper title is too long for the running head, you can set
% an abbreviated paper title here. If not, comment out.
\titlerunning{PFGS}

% TODO FINAL: Replace with your author list. 
% Include the authors' OCRID for the camera-ready version, if at all possible.
\author{
Jiaxu Wang\inst{1}$^\dagger$\orcidlink{0000-0003-1277-6896} \and
Ziyi Zhang\inst{1}$^\dagger$\orcidlink{0000-0003-2559-5269} \and
Junhao He\inst{1}\orcidlink{0009-0004-2215-1261} \and
Renjing Xu\inst{1}\textsuperscript{\Letter}\orcidlink{0000-0002-0792-8974}
% \thanks{\Letter~ Corresponding author.} 
\thanks{$\dagger$~Co-first authors; \Letter~ Corresponding authors}
}

% TODO FINAL: Replace with an abbreviated list of authors.
\authorrunning{Wang et al.}
% First names are abbreviated in the running head.
% If there are more than two authors, 'et al.' is used.

% TODO FINAL: Replace with your institution list.
\institute{$^1$ Hong Kong University of Science and Technology (GZ)\\
\email{jwang457@connect.hkust-gz.edu.cn} \\
\email{\{ziyizhang,junhaohe,renjingxu\}@hkust-gz.edu.cn}
}

\maketitle

\begin{abstract}
  Rendering high-fidelity images from sparse point clouds is still challenging. Existing learning-based approaches suffer from either hole artifacts, missing details, or expensive computations. In this paper, we propose a novel framework to render high-quality images from sparse points. This method first attempts to bridge the 3D Gaussian Splatting and point cloud rendering, which includes several cascaded modules. We first use a regressor to estimate Gaussian properties in a point-wise manner, the estimated properties are used to rasterize neural feature descriptors into 2D planes which are extracted from a multiscale extractor. The projected feature volume is gradually decoded toward the final prediction via a multiscale and progressive decoder. The whole pipeline experiences a two-stage training and is driven by our well-designed progressive and multiscale reconstruction loss. Experiments on different benchmarks show the superiority of our method in terms of rendering qualities and the necessities of our main components. \footnote[1]{Project page: \url{https://github.com/Mercerai/PFGS}}.
\end{abstract}

\section{Introduction}
\label{sec:intro}
Synthesizing photorealistic images from given colored point clouds and arbitrary camera views can be extensively applied to various fields including virtual/augmented reality \cite{jiang2024vr}, robotic navigation \cite{huang2023visual}, automatic driving \cite{zhang2022rethinking,wang2022point}, etc, which is academically named point cloud rendering. Traditional methods to render views from point clouds are based on graphics transformations. They directly project 3D point clouds into 2D planes by using the camera parameters and z-buffer rasterization. However, the sparse distribution of points leads to bleeding surfaces and hole artifacts. Even though the graphics renderer can treat points as graphical primitives and diminish these issues by color blending, rough and simple geometries of primitives entail blurred images and missing details. 

In recent years, learning-based methods \cite{rakhimov2022npbg++, hu2023trivol, hu2023point2pix} have become prevailing to address the drawbacks of conventional graphics-based approaches. These methods project points and corresponding feature descriptors to 2D planes, obtaining sparse images. Then they use UNet-like networks to complete and restore the low-quality images \cite{jena2022neural,rakhimov2022npbg++}. Nevertheless, these methods rely on 2D image restoration, which cannot guarantee 3D consistency and can result in the distinctive appearance of the same object from nearby different viewpoints. Moreover, their feature descriptors are associated with points that do not possess actual shapes and volumes in real space, thereby showing poor descriptions of local area structures. To obtain 3D-aware and consistent features, some approaches \cite{graham2017submanifold, lassner2021pulsar} transform points to 3D volumes by trilinear interpolation and use 3D convolution to extract the features. However, they lead to a significant computational burden and sacrifice the efficient representation characteristics of point clouds. For example, voxels uniformly represent scenes, whereas points are densely distributed in complex areas and sparsely distributed or even absent in empty spaces. 

To obtain efficient and consistent 3D representation, the Neural Radiance Field (NeRF) \cite{mildenhall2021nerf} has gradually been incorporated into point cloud renderers \cite{hu2023point2pix,hu2023trivol}. These methods either consider points as anchors of NeRF or transform points into NeRF-like representations, then use the physical-based volume rendering to obtain 3D-aware feature maps which are decoded into corresponding images. With the assistance of NeRF, they achieve high-quality consistencies across nearby views and plausible images. However, discrete NeRF requires heavy memory consumption resulting from storing high-dimensional feature vectors. In addition, NeRF typically requires lots of sampling points and multiple accesses to the neural network, which significantly increases the rendering time and makes it challenging to meet real-time requirements. 

More recently, 3D Gaussian Splatting (3DGS) \cite{kerbl20233d} introduces a new representation that formulates points as 3D Gaussians with learnable parameters including 3D position, color, opacity, and anisotropic covariance. 3DGS achieves more photorealistic and faster rendering than NeRF by applying Gaussian rasterization and $\alpha$-blending. Conventional 3D Gaussian learns the primitive parameters associated with an initialized point cloud from multi-view images by backpropagation, it relies on per-subject parameter optimizations for several tens minutes. It is therefore impractical to be applied to the case in which only point cloud is available due to the need for multiview images and backpropagations. 

In this work, we fill the gap between point cloud render and the 3DGS technique by introducing a novel multiscale, feature-augmented, 3DGS-based point renderer. Compared with previous learning-based point renderers, our method yields higher-quality and 3D consistent results. Compared with conventional 3DGS, our method does not necessarily require multiview images or fine-tuning procedures and can be directly employed to point modality. 

We leverage an efficient multiscale feature extractor in the proposed framework to extract features from point cloud input. This extractor considers multi-scale local geometries to extract discriminative feature descriptors. Then a regressor is used for estimating 3D Gaussian parameters in a point-wise manner. Next, we not only render images by these estimated Gaussian attributes in a multiscale way but also render the multiscale feature maps by the same Gaussian properties. Finally, the rendered multiscale images and feature maps are fused by a multiscale and recurrent UNet-like network to gradually transform into the final results. To the best knowledge of us, this is the first point cloud renderer combined with 3DGS. 

Our main contributions can be summarized in the following:
\begin{enumerate}
\item We proposed a novel lightweight approach called PFGS (Point Feature Gaussian Splatting) that bridges point cloud rendering and 3DGS. It produces higher-fidelity images from sparse point clouds than currently prevailing approaches. 
\item We proposed the point-based multiscale and feature-augmented 3D Gaussian framework, which consists of three main components including a multiscale feature extractor, a Gaussian regressor, and a multiscale and recurrent decoder, accompanied by the well-designed multiscale and progressive reconstruction losses. 
\item Extensive experiments and ablation studies on three datasets show the effectiveness and advantages of the proposed method over current works in terms of reconstruction quality. 
\end{enumerate}
\section{Related Work}

\subsection{Point Cloud Render}
Conventional point cloud rendering is implemented via graphics-based transformation \cite{ravi2020accelerating, zhou2018open3d} which simulates camera imaging processes to project points onto the 2D screens. Even though the rendering pipeline is fast and general for all scenes, it is heavily dependent on the density of point distributions. These methods produce empty holes and vacant areas when the distributions are sparse. 

Learning-based point renderers \cite{choy20194d, jena2022neural, rakhimov2022npbg++, hu2023trivol, hu2023point2pix} to some extent handle the issues caused by traditional methods, which fill the holes by leveraging multiscale or multiview feature fusions. ME \cite{choy20194d} implements a sparse Convolutional Neural Network to extract features from existing points and then calculates the features of arbitrary 3D points through ball querying. NPBG \cite{jena2022neural} enhances each point with a feature descriptor to encapsulate local geometry and appearance. Moreover, NPBG++ \cite{rakhimov2022npbg++} introduces feature aggregation from multiview based on the NPBG to complement missing parts. NPCR \cite{dai2020neural} employs a 3DCNN to generate 3D volumes from point clouds and creates multiple depth layers to synthesize images. More recently, Point2pix \cite{hu2023point2pix} and Trivol \cite{hu2023trivol} have incorporated NeRF into the point cloud rendering pipeline. They transform raw point clouds into NeRF-related representations and utilize volume rendering to access the representations to produce images. However, the above methods still cannot fill large vacant holes and only produce blurry rendering results, especially in local areas with fewer points. Even if incorporating NeRF representation is effective in generating 3D consistent feature encoding, they do not fully utilize the geometric information contained within the point clouds. In this work, we combine point rendering with a novel 3D representation, i.e. 3DGS, and implement multiscale and progressive feature decoding to obtain fine-grained and 3D-consistent predictions.

\subsection{Neural-based View Synthesis}
Unlike traditional explicit representations \cite{qi2017pointnet++, wang2019mvpnet, wang2018pixel2mesh, qi2016volumetric}, implicit representations that encode scenes with the assistance of neural networks have been developed in recent years \cite{lombardi2019neural, sitzmann2019deepvoxels, xiang2021neutex}. Among them, the Neural Radiance Field (NeRF)\cite{mildenhall2021nerf} attracts the most attention in both industries and research. NeRF is a continuous method that implicitly represents a 3D scene employing an MLP and synthesizing images of novel view by volume rendering. Recent studies have employed NeRF for various purposes, including dynamic reconstruction \cite{yan2023nerf, cao2023hexplane, wu20234d}, physical reasoning \cite{wang2024physical}, generative models \cite{metzer2023latent, mypg, wang2022clip}, and fast rendering\cite{fridovich2022plenoxels, muller2022instant,wang2023f2}. To achieve various goals, some studies attempted to combine the implicit NeRF with explicit primitives. For example, NeuMesh \cite{yang2022neumesh} regards mesh as a neural scaffold to store NeRF and achieve neural editing. PointNeRF \cite{xu2022point} is the first to incorporate point clouds into NeRF reconstruction to obtain better rendering qualities and faster speed. Ref-NeuS \cite{ge2023ref} implements generalizable neural surface reconstruction. GPF \cite{wang2024learning} explicitly models the visibility by using the geometries in the point cloud. PointNeRF++ \cite{sun2023pointnerf++} solves point cloud sparsity by aggregating multiscale point clouds on sparse voxel grids at different resolutions. 

It can be seen that many approaches we list above incorporate points into NeRF. However, NeRF-like volume rendering does not fully explore the rendering capability of the point cloud. In contrast to NeRF, 3DGS \cite{kerbl20233d} proposed the 3D Gaussian Splatting to realize a remarkable differentiable rendering speed and yield high-quality 2D images. In this work, we first bridge point cloud rendering and 3DGS to generate higher-fidelity images.

\section{Methodology}
Given a colored point cloud $\mathbf{P}=\{\mathbf{p}_k, \mathbf{c}_k \}$, we aim to synthesize photorealistic images from any viewpoints defined by the camera intrinsic $K$ and poses $P$ are provided. The $\mathbf{p}_k \in R^3$ and $\mathbf{c}_k \in R^3$ are the point coordinates and colors respectively. The rendering process can be illustrated as $I_v = Re(\mathbf{P}|K_v, P_v)$ where $Re$ is the rendering function that can be implemented by either graphics-based or learning-based methods as we introduced above. In this work, it is implemented by our proposed multiscale feature-based 3DGS rendering pipeline that contains a multiscale feature extractor, a Gaussian regressor, a feature-based Gaussian rendering module, and a multiscale recurrent decoder. All modules are implemented with small networks.
In the following of this section, we sequentially introduce the aforementioned components. 

\begin{figure}[t]
    \centering
    \includegraphics[width=\textwidth]{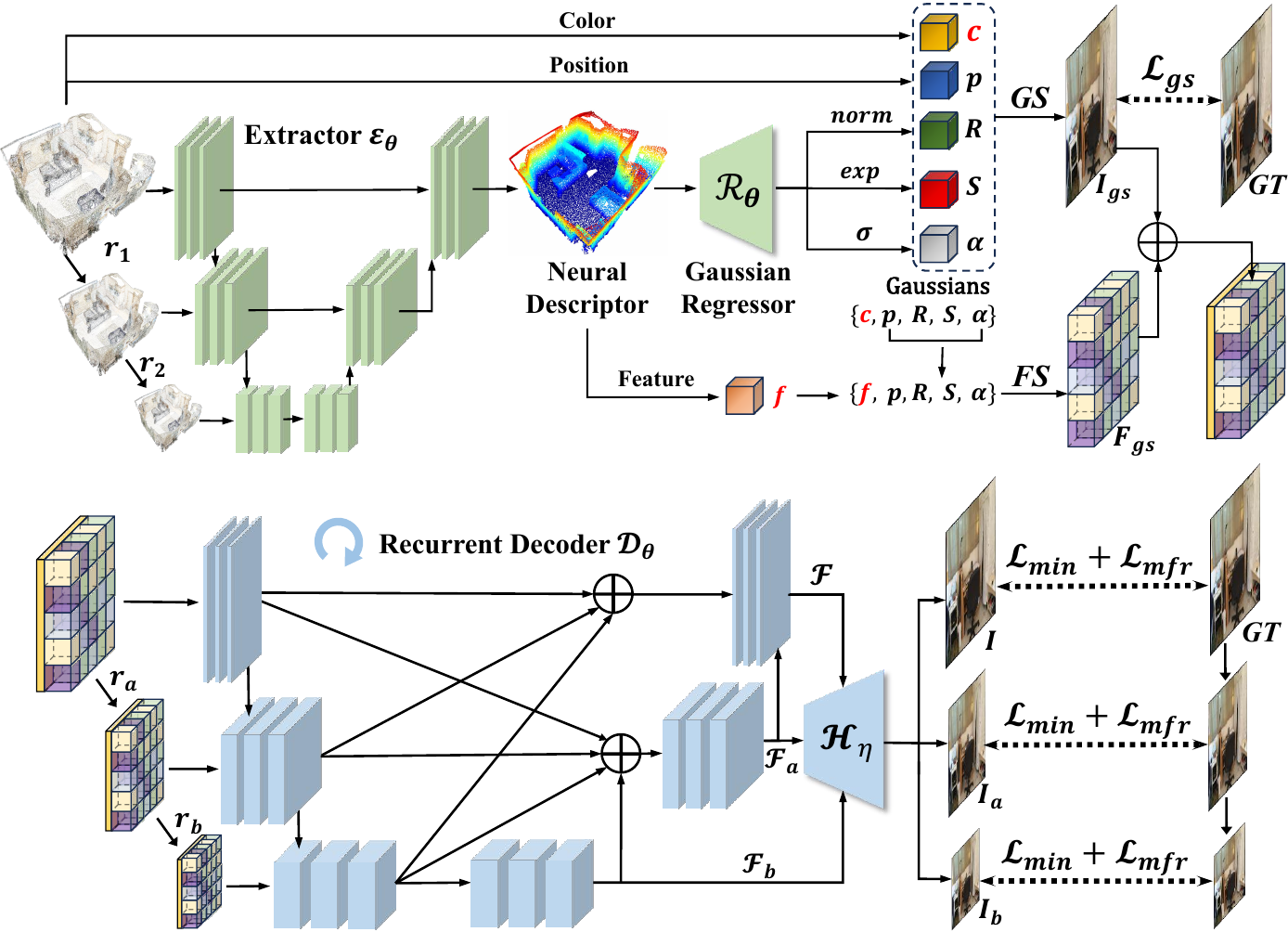}
    \caption{The main pipeline of the proposed approach. The above panel includes the multiscale feature extractor and Gaussian regressor. Both Gaussian and features are splatted to the 2D plane and concatenated and fed to the progressive and multiscale feature decoding module described in the low panel. }
    \label{fig: main pipeline}
    \vspace{-0.5cm}
\end{figure}

\subsection{Preliminaries}
3DGS is an explicit representation method for 3D scenes, which parameterizes the scene as a series of 3D Gaussian primitives. A 3D Gaussian is defined by a full 3D covariance matrix $\Sigma$, its center point $x$, and the spherical harmonic ($SH$). The mean value of the Gaussian is defined as:
\begin{equation}
    G(x)=e^{-\frac{1}{2}(x)^{T}\Sigma^{-1}(x)}
\end{equation}
To enable optimization via backpropagation, the covariance matrix could be decomposed into a rotation matrix ($R$) and a scaling matrix ($S$):
\begin{equation}
    \Sigma=RSS^{T}R^{T}
\end{equation}
Given the camera trajectory, the projection of the 3D Gaussians to the 2D image plane can be characterized by a view transform matrix ($W$) and the Jacobian of the affine approximation of the projective transformation ($J$), as in:
\begin{equation}
    \Sigma^{'}=JW\Sigma W^{T}J^{T}
\end{equation}
where the $\Sigma^{'}$ is the covariance matrix in 2D image planes. Thus, the $\alpha$-blend of  $\mathcal{N}$ ordered points overlapping a pixel is utilized to compute the final color $C$ of the pixel:
\begin{equation}
    C = \sum_{i\in \mathcal{N}}c_i\alpha_i\prod_{j=1}^{i-1}(1-\alpha_j)
    \vspace{-1mm}
\end{equation}
where $c_{i}$ and $\alpha_{i}$ denote the color and density of the pixel with corresponding Gaussian parameters.

In summary, the above parameters can be characterized by the following attributes: $x$ denotes the position of a primitive where $x \in R^{3}$. The rotation matrix is characterized by a quaternion $q \in R^{4}$. The scale factor is defined by the anisotropy stretching $s \in R^{3}$. The opacity factor $\alpha \in R^{1} \in [0, 1]$. The 2D opacity $\alpha \in [0,1]$ is computed by $\alpha_i(x) = o_iexp(-\frac{1}{2}(x-\mu_i)^T{\Sigma}_{i}^{T}(x-\mu_i))$ where the $\mu$ and $variance$ are the 2D-projected mean and variance of 3D Gaussians. And the color is represented by $SH$.

\subsection{Gaussian Feature Prediction with Multiscale Feature Extraction}
We first extract features for each point as their neural descriptors. We use a multi-input, single-output UNet-like architecture as the extraction network to encode points with different scales and capture features across different spatial distances. The architecture is mainly made of several PV Convolution modules \cite{liu2019point}, which is a highly efficient and commonly used point-based feature integration operation. We downsample the original point cloud with decreasing rates and concatenate the downsampled point clouds with different levels of feature maps as extra inputs. 
\begin{equation}
\begin{aligned}
    f_{1:k} = \mathcal{E}_{\theta}((p_{1:k},c_{1:k}), (p_{1:k},c_{1:k})_{\downarrow r_1}, (p_{1:k},c_{1:k})_{\downarrow r_2}) 
    \label{eq: extractor}
\end{aligned}
\end{equation}
The $\mathcal{E}_{\theta}$ refers to the extractor network, $p,c$ denote point coordinates and colors. The down arrow represents a uniform downsampling operation and $r$ denotes the sampling rate where $r_1 > r_2$. The output $f_{1:k}$ are point neural descriptors. 
This operation enables the extractor to recognize features of different scales and receptive fields. The detailed architecture and parameter setting are shown in our Appendix. The single output feature vector is considered as the descriptors for all points. Afterward, the neural descriptors, the coordinates, as well as colors are fed to another network to predict the Gaussian properties in a point-wise manner.
\begin{equation}
\begin{aligned}
    &F_k = \mathcal{R}_{\theta}(f_k, p_k, c_k) \\
    &R, S, \alpha = norm(\mathcal{H}_r(F_k)), exp(\mathcal{H}_s(F_k)), \sigma(\mathcal{H}_o(F_k)) 
    \label{eq: regressor}
\end{aligned}
\end{equation}
where $\mathcal{R}_{\theta}$ represents the main network of the regressor, it maps multiple inputs to a single unified feature vector $F_k$. The $F_k$ is fed to three individual prediction heads ($\mathcal{H}_r$, $\mathcal{H}_s$, $\mathcal{H}_o$), each of them is composed of two 1D Convolution layers with 1 kernel size, and activated by different functions to obtain corresponding Gaussian properties, namely rotation quaternion ($R$), scale factor ($S$) and opacity ($\alpha$). The color property $c$ remains that of the original points.

After the Gaussian attributes are obtained, one can render corresponding images from given camera viewpoints by Gaussian rasterization. We not only render the images by blending point colors on 2D planes but also render feature planes by replacing colors with the neural descriptors ($f_k$). 
\begin{equation}
    I_{gs}, F_{gs} = Re(R, S, \alpha, p_{1:k}, c_{1:k}|K,P), Re(R, S, \alpha, p_{1:k}, f_{1:k}|K,P)
    \label{eq: render feature map}
\end{equation}
where $I_{gs}$, $F_{gs}$ are corresponding image and feature map under the camera intrinsic $K$ and pose $P$. The $I_{gs}$ can be considered a good result of point rendering when the number of points is sufficient. 

However, when the point sparsely distributes over the space, the representation capability of corresponding Gaussians is constrained, especially in complex local areas, resulting in poor-quality rendering results. On the contrary, the neural descriptors contain rich information to describe local structures with various scales. We therefore utilize the $F_{gs}$ to be gradually transformed to the final results in a multiscale and recurrent manner. The $F_{gs}$ is 3D-consistently rasterized with the same Gaussian parameters in Eq.~\ref{eq: render feature map}, thus it precisely corresponds to the $I_{gs}$ in pixel-level. In other words, $F_{gs}$ can be regarded as a complement to $I_{gs}$. Hence we concatenate $I_{gs}$ and $F_{gs}$ to be the input of the next module.  

\subsection{Multiscale and Progressive Feature Decoding}
After we render the feature volume $F_{gs}$ associated with the provided camera viewpoint, we concatenate the $I_{gs}$ and $F_{gs}$ as input of the decoder. The decoder is a multi-input and multi-output UNet-like architecture, which aims to gradually refine and decode the input features and transform them into the target view images. We downsample the input two times with the rate of $r_a$ and $r_b$ and fuse them with the next two low-resolution layers of the UNet. There are interlaced connections between layers of different levels in the network, as stated in the main Fig.~\ref{fig: main pipeline}. 
\begin{equation}
    \mathcal{F}, \mathcal{F}_a, \mathcal{F}_b=\mathcal{D}_\theta((F_{gs} \oplus I_{gs}), (F_{gs} \oplus I_{gs})_{\downarrow r_a}, (F_{gs} \oplus I_{gs})_{\downarrow r_b})
    \label{eq: MIMO UNet}
\end{equation}
In the above equation, $\mathcal{D}_\theta$ indicates the multiscale decoder. $\downarrow r_a$ and $\downarrow r_b$ are two downsampling operations. $\mathcal{F}, \mathcal{F}_a, \mathcal{F}_b$ refer to the feature output of different levels of layers in which $\mathcal{F}$ denotes the full resolution feature output while $\mathcal{F}$ with subscripts $a$ and $b$ represent the lower resolution feature outputs. We apply a prediction head (marked as $\mathcal{H}_\eta$) composed of two convolutional layers and a ReLU activation to transform these features into the target images with different scales. 

More importantly, we recurrently run the decoder network twice to refine the feature maps and the predicted images. The following experiments show the positive influence of such recurrent refinement on restoring the missing details due to the sparse point distribution. The input for the second loop is related to the output of the first loop. The prediction after the first loop can be described as $I^l, I^l_a, I^l_b = \mathcal{H}_\eta(\mathcal{F}), \mathcal{H}_\eta(\mathcal{F}_a), \mathcal{H}_\eta(\mathcal{F}_b)$ where $l=1$ refers to the loop number. Likewise, the second loop is accordingly explained as:
\begin{equation}
    I^{l=2}, I^{l=2}_a, I^{l=2}_b=\mathcal{H}_\eta \star \mathcal{D}_\theta((\mathcal{F} \oplus I^{l=1}), (\mathcal{F}_{a} \oplus I^{l=1}_a),  (\mathcal{F}_{b} \oplus I^{l=1}_b))
    \label{eq: MIMO UNet loop 2}
\end{equation}
in which $\mathcal{H}_\eta \star \mathcal{D}_\theta$ represents sequentially perform the two operations. We show the detailed architecture of the decoder in the Appendix. After we obtain the hierarchically multiscale predictions, we regard the full resolution result of the second loop $I^{l=2}$ as the final prediction. However, we leverage all predictions to optimize the full models, which we introduce in the next section. 

\subsection{Training Strategies}
The proposed framework incorporates 3D-aware features into multiscale and progressive decoding by using 3D Gaussian Splatting. The integration of feature volumes under certain camera settings mainly relies on the Gaussian properties that are used for projections. Therefore, we first pretrain the feature extractor and Gaussian regressor in order to obtain a relatively reasonable rendering function with 8 epochs. In the first training stage, we employ the $L1$ rendering loss between the Gaussian predicted images and labels, i.e. $I_{gs}$ in Eq.~\ref{eq: render feature map}. 
\begin{equation}
\mathcal{L}_{gs}=\sum_{\mathbf{u} \in \Omega}\frac{1}{\Omega}||I_{gs}(\mathbf{u})-I_{gt}(\mathbf{u})||_1^1
    \label{eq: gs rendering loss}
\end{equation}
where $\mathbf{u}=\{u,v\}$ and $\Omega$ is the pixel domain. After the first pretraining, the regressor can yield plausible Gaussian rendering parameters, thus the rendered feature maps provide a certain and view-dependent description of the scene. Based on this, we combine the rest modules with these trained modules to carry out joint optimization. According to the above illustration, the final outputs of the framework include two levels, each comprising three images of different scales. We leverage all of the outputs to optimize the model by using progressively multiscale image loss.
\begin{equation}
\mathcal{L}_{mim}=\sum_{i=l}^2 \sum_{s=r}^3 \sum_{\mathbf{u} \in \Omega} w(l)\frac{1}{\Omega}||I_s^i(\mathbf{u})-{I_{gt}}_{\downarrow r_s}(\mathbf{u})||_1^1
    \label{eq: multi scale rendering loss}
\end{equation}
in which $l$ refers to the loop number in the progressive decoder. $s$ denotes the scale number in each output of a loop and $\downarrow r_s$ is $s$' corresponding downsampling operations. $w(l)$ is a simple weighting function related to $l$. Here we set $w(l)=0.75$ when $l=1$ otherwise $w(l)=1$. In addition, we use the extra progressively multiscale frequency reconstruction loss to assist in restoring the high-frequency components, it is effective in reconstructing sharper edges and boundaries of images. 
\begin{equation}
    \mathcal{L}_{mfr}=\sum_{i=l}^2 \sum_{s=r}^3 \sum_{\mathbf{u} \in \Omega} w(l)\frac{1}{\Omega}||FFT(I_s^i(\mathbf{u}))-FFT({I_{gt}}_{\downarrow r_s}(\mathbf{u}))||_1^1
    \label{eq: multi scale frequency loss}
\end{equation}
in which the $FFT$ represents the fast Fourier transform (FFT) that transfers the image signal to the frequency domain. The remaining symbols retain the same meanings as in the previous context.
The total loss function can be defined as follows:
\begin{equation}
    \mathcal{L}_{total}=\gamma_{1} \mathcal{L}_{gs} + \gamma_2\mathcal{L}_{mim} + \gamma_3\mathcal{L}_{mfr}
    \label{eq: total loss}
\end{equation}
The $\gamma_1$, $\gamma_2$, and $\gamma_3$ respectively control weights of
these losses. In our implementation, we set $\gamma_1=0.75$, $\gamma_2=1$, and $\gamma_3=0.25$ respectively. 

\section{Experiments}
\subsection{Experiments setup}
\textbf{Datasets.} We conduct extensive experiments to evaluate the effectiveness of our approach on three different datasets, including the indoor scene dataset ScanNet \cite{dai2017scannet}, objects dataset DTU \cite{jensen2014large}, and human body dataset THuman2.0 \cite{tao2021function4d}. The ScanNet dataset includes more than 2.5 million images at different views and with different qualities in over 1500 scenes. The point clouds are constructed from RGBD images. Following the Trivol, \cite{hu2023trivol} we select the first 1200 scenes as the training set and the rest of the 300+ scenes as the test set. In addition, we use the preprocessed sparse point cloud in the dataset to test our model. For the DTU dataset, we transformed the point clouds of objects from their depth maps and images and 
we downsampled the point cloud by a factor of 0.3 to keep their point clouds distributing sparsely. In addition, we follow the MVSNeRF \cite{chen2021mvsnerf} to split the train and test groups of scans. The THuman2.0 dataset contains 500 high-quality human scans captured by a dense DLSR rig. For each scan, we render 36 views based on the provided 3D model and sparse sample points on the surface of the model. We keep the number of points for each scan at around 80,000 for their sparsity. 
% We define 
% The first 80$\%$ is for the training set, and the last 20$\%$ is for the test set.
We randomly select 75$\%$ as the training set and the remaining 25$\%$ as the test set.

\noindent \textbf{Baselines and Metrics.} 
We compare the proposed method with the currently prevailing point cloud rendering approaches to demonstrate the effectiveness of our framework. The baselines include a pure graphics-based renderer, learning-based renderers NPBG \cite{jena2022neural} and NPBG++ \cite{rakhimov2022npbg++}, and the NeRF-based renderer Trivol \cite{hu2023trivol}. As for the quality, we adopt PSNR, SSIM \cite{wang2004image}, and LPIPS \cite{zhang2018unreasonable}. These metrics overall determine the reconstruction accuracy and quality of these methods. 

\begin{table}[t]
\caption{Quantitative Results of comparisons between our method and different baselines under three datasets.}
\vspace{-2mm}
\renewcommand{\arraystretch}{1.2}
\setlength\tabcolsep{0.85mm}{
\centering
\setlength\tabcolsep{0.86mm}{
\begin{tabular}{l|ccc|ccc|ccc}
\hline\hline
&\multicolumn{3}{c|}{ScanNet} &\multicolumn{3}{c|}{DTU} & \multicolumn{3}{c}{THuman2.0}  \\ 
&PSNR↑ &SSIM↑ &LPIPS↓ &PSNR↑ &SSIM↑ &LPIPS↓ &PSNR↑ &SSIM↑ &LPIPS↓\\
\hline
Graphics&13.62 &0.528 &0.779 &12.15 &0.525 &0.682 &20.26 &0.905 &0.337\\
NPBG&15.09 &0.592 &0.625 &13.52 &0.703 &0.514 &19.77 &0.915 &0.112\\
NPBG++&16.81 &0.671 &0.585 &\underline{22.32} &\underline{0.833} &\underline{0.327} &\underline{26.81} &\underline{0.952} &0.062\\
Trivol&\underline{18.56} &\underline{0.734} &\underline{0.473} &19.25 &0.592 &0.518 &25.97 &0.935 &\underline{0.059}\\
Ours&\textbf{19.86} &\textbf{0.758} &\textbf{0.452} &\textbf{25.44} &\textbf{0.901} &\textbf{0.164} &\textbf{34.74} &\textbf{0.983} &\textbf{0.009}\\
\hline\hline
\end{tabular}}}
\label{Tab : Comparison}
\vspace{-2mm}
\end{table}

\begin{figure}[t]
    \centering
    \includegraphics[width=\textwidth]{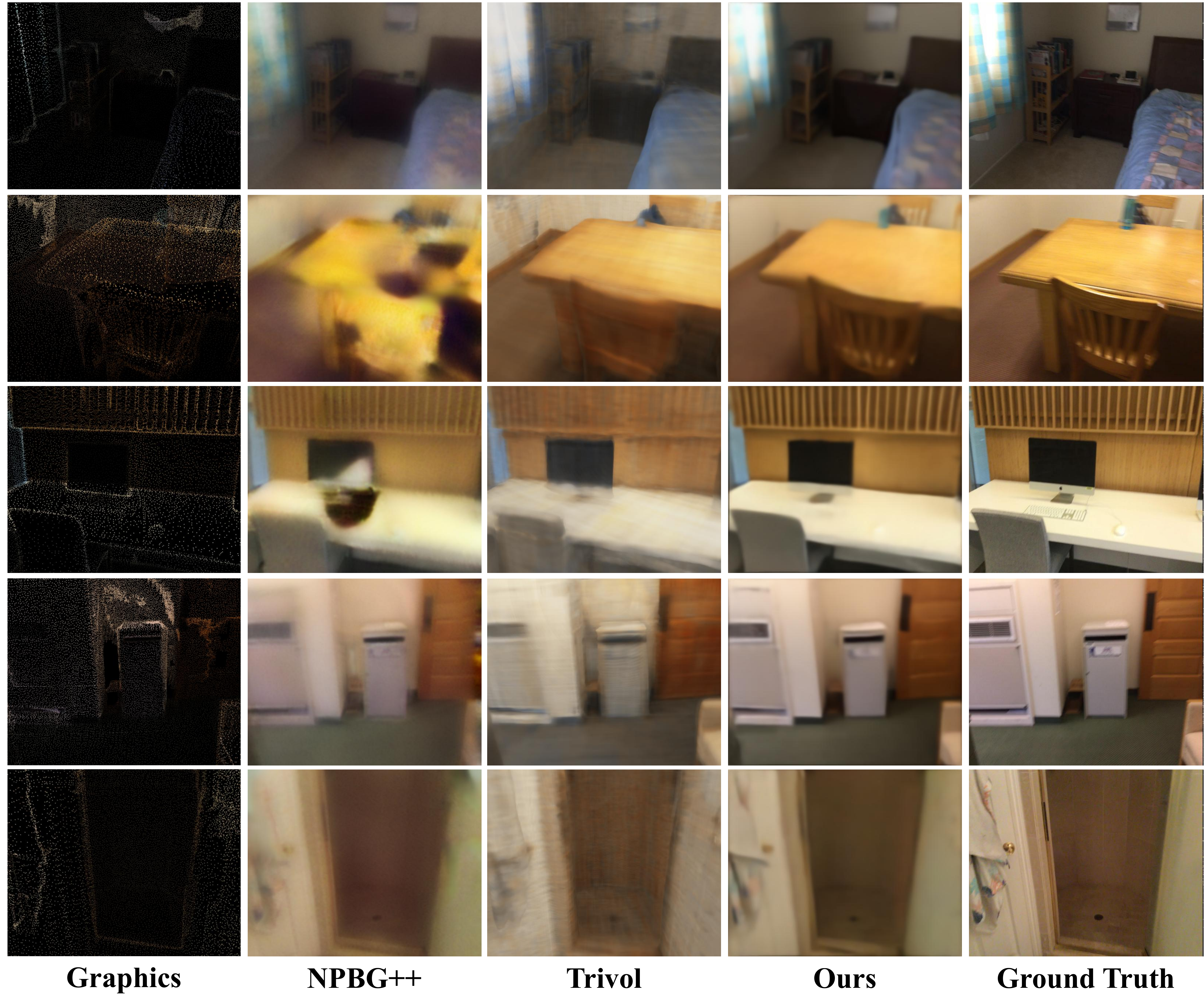}
    \vspace{-5mm}
    \caption{Qualitative comparisons between ours and other point cloud renderers on the ScanNet dataset.}
    \label{fig: scannet}
    \vspace{-7mm}
\end{figure}

\begin{figure}[t]
    \centering
    \includegraphics[width=\textwidth]{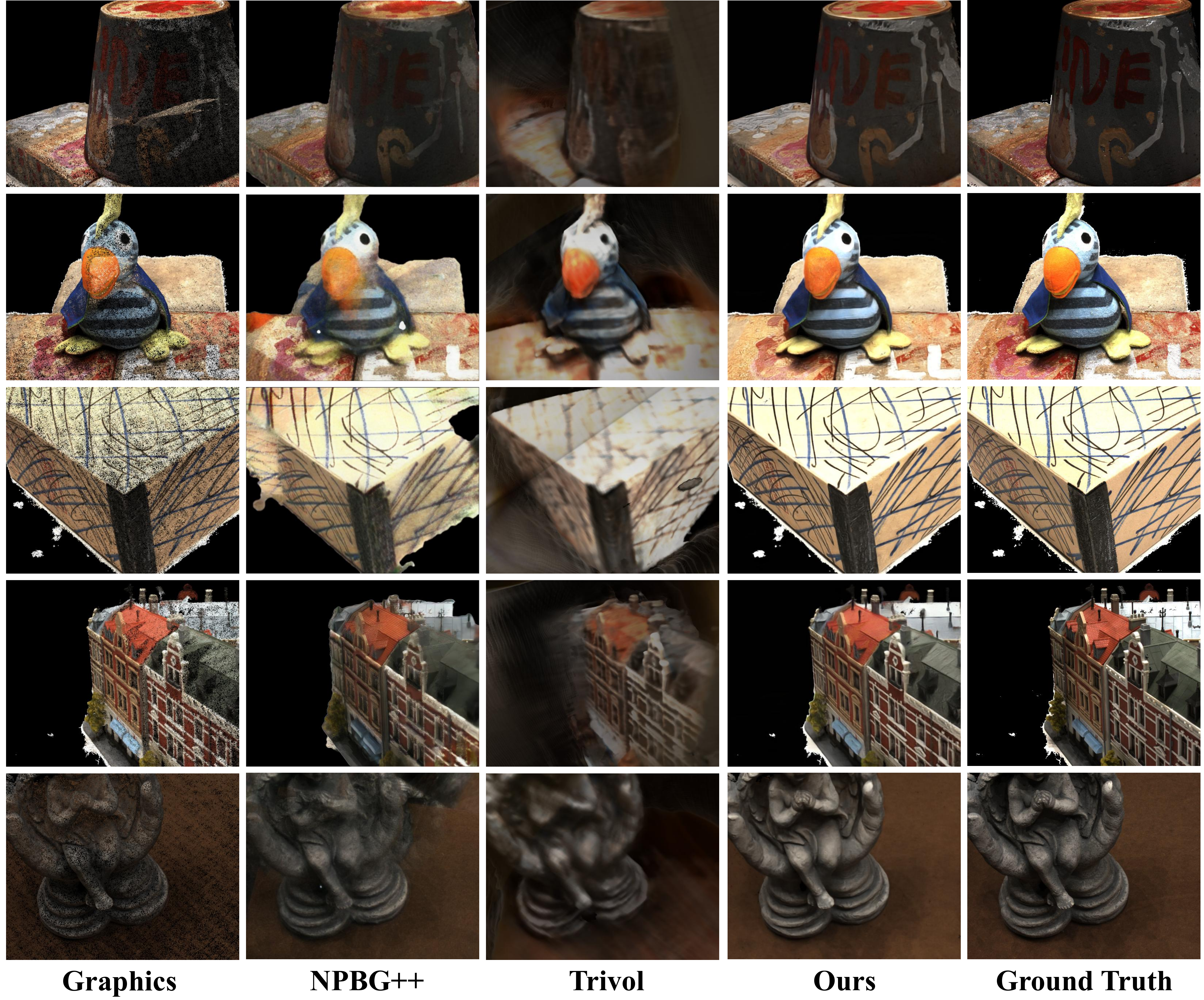}
    \vspace{-5mm}
    \caption{Qualitative comparisons between ours and other point cloud rendering methods on the DTU dataset.}
    \label{fig: dtu}
    \vspace{-5mm}
\end{figure}

\begin{figure}[t]
    \centering
    \includegraphics[width=\textwidth]{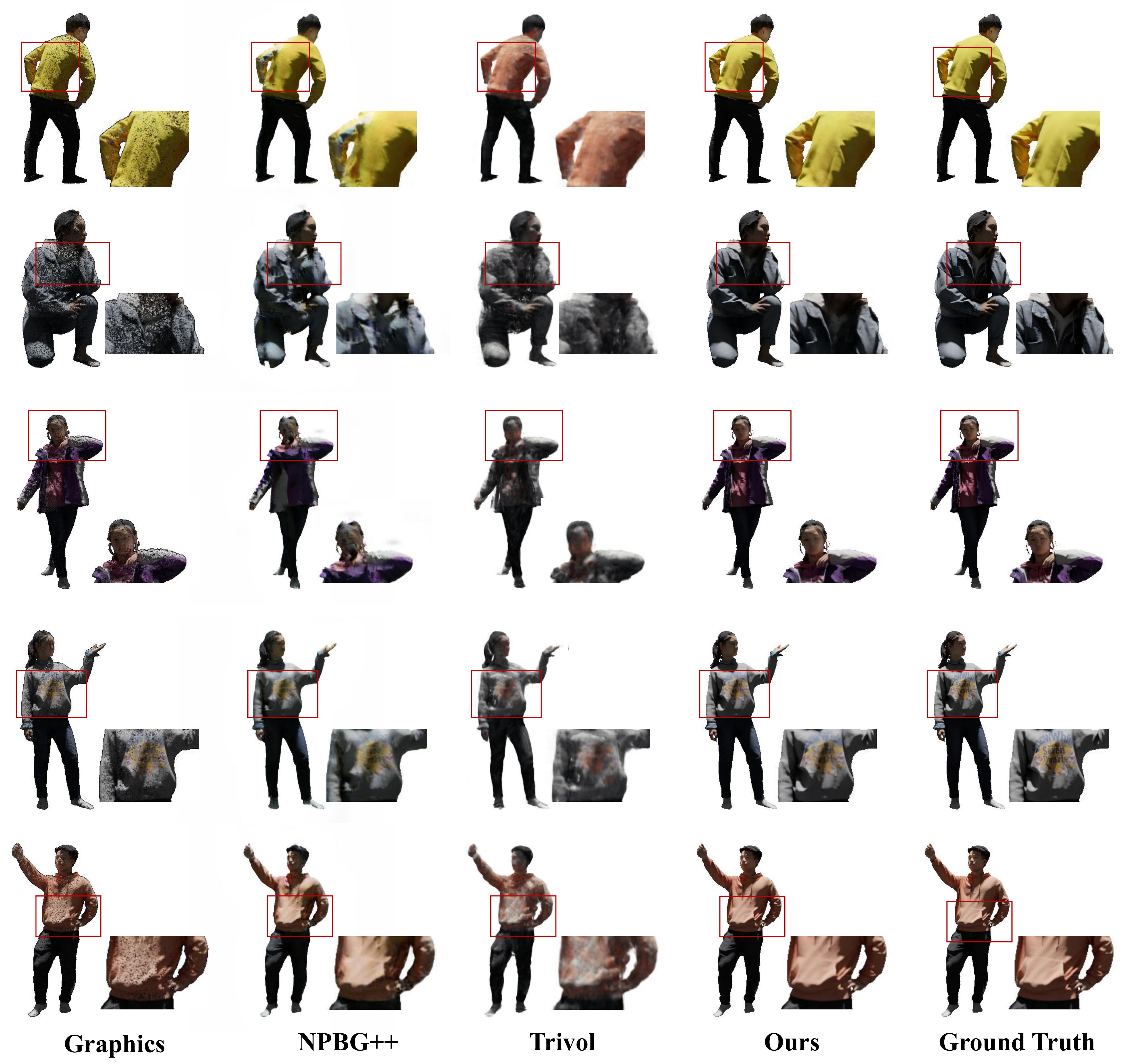}
    \vspace{-5mm}
    \caption{Qualitative comparisons between ours and other point cloud renderers on the THuman2.0 dataset.}
    \label{fig: thuman2}
    \vspace{-8mm}
\end{figure}

\subsection{Comparisons With Other Point Cloud Renderers.}
Fig. \ref{fig: scannet}, \ref{fig: dtu}, and \ref{fig: thuman2} show the results of qualitative comparisons on the ScanNet, DTU, and THuman2.0 datasets respectively. Table.~\ref{Tab : Comparison} reports all the metrics on the three datasets of the baselines and ours. The best result is bold, and the second-best result is underlined in the Table. 

\noindent \textbf{ScanNet.} The problem of uneven distribution and missing points exists in the real scene-level point cloud. As shown in the first column in Fig. \ref{fig: scannet}, traditional graphical rendering leads to a large number of holes in the rendering result. When the point cloud distribution is uneven, the point cloud of nearby objects is relatively sparse, and the 2D-CNN of NPBG++ pays more attention to the features of distant objects, ignoring the 3D information in the point cloud, resulting that nearby objects are covered in the rendering result. Our method uses 3DGS for feature projection, which better integrates 3D context information. There are a large number of artifacts in the results of NPBG++ and Trivol. Our multi-scale feature extraction network excels at capturing features across various levels of detail, effectively decoding these projected features to enhance the rendering of lighting effects and details significantly. 

\noindent \textbf{DTU.} As we stated in the Dataset description part, we maintain points distributed sparsely on the surface of the objects. Under this circumstance, the graphics-based transformation demonstrates more serious bleeding surfaces. Trivol always partially reconstructs the whole object because it depends on the fully implicit neural representation. This setting is more likely to be suited to class-specific point rendering, as they claimed in the original paper. In contrast, DTU includes many different classes that obstacle Trivol to learning a unified feature transform underlying them. NPBG++ and our method deliver better results than Trivol because they leverage the original color and structure of point clouds without class-specific assumptions. Our approach yields the clearest predictions thanks to our more reasonable feature projection and decoding. 

\noindent \textbf{THuman2.0} The Graphics-based method reconstructs the main structures of the human body while inevitably seeing densely packed holes. The other two baselines generate similar results that are much better than those of the traditional methods. The difference is that Trivol fails to faithfully recover some cloth colors, which is because it is a fully implicit method and thereby cannot restore colors that are never seen in its train set. NPBG++ and ours directly utilize points' colors thus avoiding this domain generalization issue. Moreover, our method restores more details such as human facial expressions and clothes' folds.  
\vspace{-3mm}

\begin{table}[t]
\vspace{-2mm}
\caption{Quantitative Results of ablation studies under various variants on the ScanNet dataset. The meaning of each abbreviation can be found in the text.}
\renewcommand{\arraystretch}{1.2}
\centering
\setlength\tabcolsep{1mm}{
\begin{tabular}{lcccccccc}
\hline\hline
  & w/o ProL & w/o MSL   & w/o MSf  & w/o Lmfr & w/o 2tr & recur1 & recur3 & Full \\ \hline
PSNR↑    & 19.19 & 19.37 & 18.27  & 19.42 & 15.92 & 19.19 &19.77 & \textbf{19.86}  \\
SSIM↑    & 0.725 & 0.733 & 0.704  & 0.740 & 0.696 & 0.729 &0.760 & \textbf{0.758}  \\
LPIPS↓   & 0.498 & 0.509 & 0.650  & 0.478 & 0.689 & 0.516 &0.478 & \textbf{0.452}  \\
\hline\hline
\end{tabular}}
\label{Tab: Ablation}
\end{table}

\begin{figure}
    \centering
    \includegraphics[width=\textwidth]{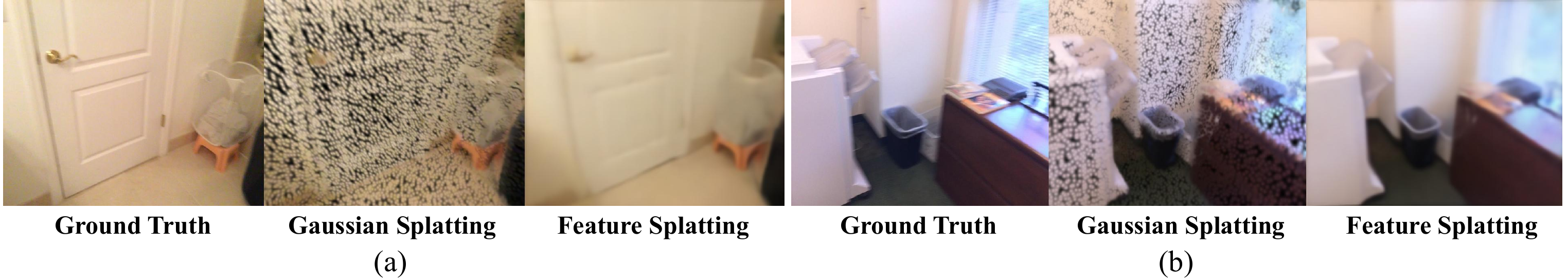}
    \vspace{-6mm}
    \caption{Qualitative Comparisons of the renderings produced by the Gaussian splatting after the first stage and the feature splatting after the whole framework.}
    \label{fig: ablations of gs and fs}
    \vspace{-4mm}
\end{figure}

\subsection{Ablation studies.}
% \vspace{-2mm}
\subsubsection{Ablations of main components.}
In this subsection, we evaluate the effectiveness of all main constituents proposed in our whole framework. We conduct experiments with different control variables set and report their performance on the ScanNet dataset in Table~\ref{Tab: Ablation}. In the table, $w/o~Pro L$ refers to the version that removes all the progressive loss items, i.e. removes the first summation in Eq.~\ref{eq: multi scale rendering loss} and Eq.~\ref{eq: multi scale frequency loss}. $w/o~MS L$ denotes the version that removes all the multiscale loss items that are the second summation in Eq.~\ref{eq: multi scale rendering loss} and \ref{eq: multi scale frequency loss}. The $recur1$ illustrates that we discard the recurrent structure from the feature decoder and only utilize a single multi-input and multi-output UNet. Besides, the $w/o~MSf$ describes a variant of the decoder which adopts a normal single scale UNet. In other words, this variant receives a single feature map as input and outputs a single prediction with full resolutions. Therefore, we only maintain the progressive losses and structures but discard all multiscale components for this version. Furthermore, we evaluate when the number of loops equals to 3, i.e. $l=3$ in these equations. This is marked as $recur3$ in the Table. The results show that continuously increasing the number of iterations does not further improve the model's performance; instead, it may lead to a slight decline. Last, the $w/o~Lmfr$ represents the version of removing the progressively multiscale frequency reconstruction loss (Eq.~\ref{eq: multi scale frequency loss}). 

It is observed that the full model delivers the most satisfactory rendering results across all metrics while the rest variants experience a decline to some extent. This indicates that both progressive and multi-scale features have a positive impact and contribute to the final promising performance of the model. Therefore, the entire design of this method is consistently characterized by progressive and multi-scale aspects. Moreover, the alignment of predictions and labels in the frequency domain progressively and in a multiscale manner makes prediction clearer and sharper.

\subsubsection{Analysis of Gaussian and Feature Splatting.}
The proposed approach relies on the two-stage training scheme because the 3D-consistent feature should be projected and integrated into certain 2D planes given camera viewpoints. The projection of features is dependent on the 3D Gaussian rasterization and further reliant on the Gaussian parameters associated with each point which are estimated by Eq.~\ref{eq: regressor}. The significance of training the Gaussian regression network in the first training stage lies in laying a solid foundation for feature splatting. Even though the 3D Gaussian Splatting could also produce plausible images, namely the $I_{gs}$ in Eq.~\ref{eq: render feature map}, the results are still severely influenced by points that are either locally or globally sparse, thereby missing details and showing holes. It can be seen in Fig.~\ref{fig: ablations of gs and fs} that these Gaussian splatting results are indeed better than the pure graphics-based method due to the representation capabilities of 3DGS, but they still need to be improved further.

However, the results of feature splatting contain rich semantic information and local geometry cues, we consider them a good representation of restoring high-quality images and filling up blank areas. From Fig.~\ref{fig: ablations of gs and fs} we can observe the capability of our method to generate and complete areas with voids and regions with sparse point clouds. We additionally evaluate an experiment in which we collaboratively train all modules in the framework rather than conducting the two-stage training and the results on the ScanNet dataset and we add this additional result into Table.~\ref{Tab: Ablation}. $w/o~2tr$ means that we abandon the first training stage and jointly train all modules together. In this setting, we observe that the model completely collapses and sees a very fast degradation in performance.  

\begin{table}[t]
\caption{Ablation studies of the effect of different numbers of points on performance.}
\renewcommand{\arraystretch}{1.2}
\centering
\setlength\tabcolsep{4.2mm}{
\begin{tabular}{lcccccccc}
\hline\hline
Train Points  & 50k &  50k  & 50k  & 120k & 120k  & 120k   \\ \hline
Test Points  & 50k &  80k  & 120k &  50k &  80k  & 120k  \\ \hline
Trivol & 24.05 & 25.43 & 26.48  & 24.71 & 25.96 & 26.86  \\
Ours  & 31.29 & 32.86 & 35.95  & 31.60 & 34.17 & 39.18   \\
\hline\hline
\end{tabular}}
\label{Tab: number of points}
\vspace{-4mm}
\end{table}
\subsubsection{Analysis of The Number of Points.} We additionally analyze and evaluate how different training and test numbers of points affect the performance. On the THuman2.0 dataset, we set different training and test pairs to show the performance of our method and Trivol respectively, which includes 50,000 and 120,000 points used for training and 50,000, 80,000, and 120,000 points used for testing. The results are listed in Table~\ref{Tab: number of points}. It is noted that Trivol is essentially unaffected by the number of points where our method sees significant fluctuations with changes in the number of points. However, our method exhibits very high performance when point clouds are abundant while the Trivol method has a relatively low upper limit, which is due to its fully implicit representation and lack of utilization of point cloud prior. More discussion can be seen in the Appendix.

\section{Discussion}
\textbf{Conclusion.} This work proposes a novel point cloud rendering method called PFGS, that can render high-fidelity and photorealistic images from colored point clouds with arbitrary camera viewpoints. To address the shortcomings of current learning-based point cloud renderers such as slow, blurry, and inconsistent, we introduce the 3D Gaussian Splatting to point cloud rendering by several well-designed modules. We supervise the regressor to estimate Gaussian parameters in a point-wise way, and then we utilize the estimated Gaussian parameters to rasterize point-wise neural features that are extracted by a multiscale feature decoder. Furthermore, the projected feature volume would be gradually decoded to the final prediction by our progressive and multiscale decoder. The whole architecture is supervised by our finely-designed progressive and multiscale reconstruction losses.  We carry out experiments on three datasets with different levels of objects including indoors, objects, and humans and the results show our method outperforms all baselines. The ablation studies also illustrate the significance of all main components in the approach. 

\noindent\textbf{Limitation.} Even though the proposed method can produce photorealistic images from sparse and uneven point clouds, it still yields unreal details when the given point clouds are not precise enough. Future works can incorporate a large generative refiner trained on a large dataset to slightly refine the original point cloud to make points closer to the actual surface of objects. 
% Moreover, although the FGS can achieve real-time rendering and is faster than NeRF-based renderers, it is still slower than the graphics-based model. Future work can focus on acceleration.

\clearpage

\bibliographystyle{splncs04}
\bibliography{main}

\clearpage 

\appendix
{\huge{\centerline{\textbf{Appendix}}} } 

\section{More Details of the Proposed Method}
\subsection{Implementation Details}
We train our model on a single NVIDIA RTX3090 GPU using Adam optimizer and set the initial learning rate as 0.0001. We trained Stage I for 10K iterations, which took about 5 hours to converge. Stage I only contains multiscale feature extraction and Gaussian regression and outputs Gaussian splatting image and feature splatting map. Then, we trained Stage II for 60K iterations, which took about 36 hours to converge. Stage II includes the full model.
\subsection{Multiscale Feature Extraction and Gaussian Regressor}
\begin{figure}
\vspace{-4mm}
    \centering
    \includegraphics[width=\textwidth]{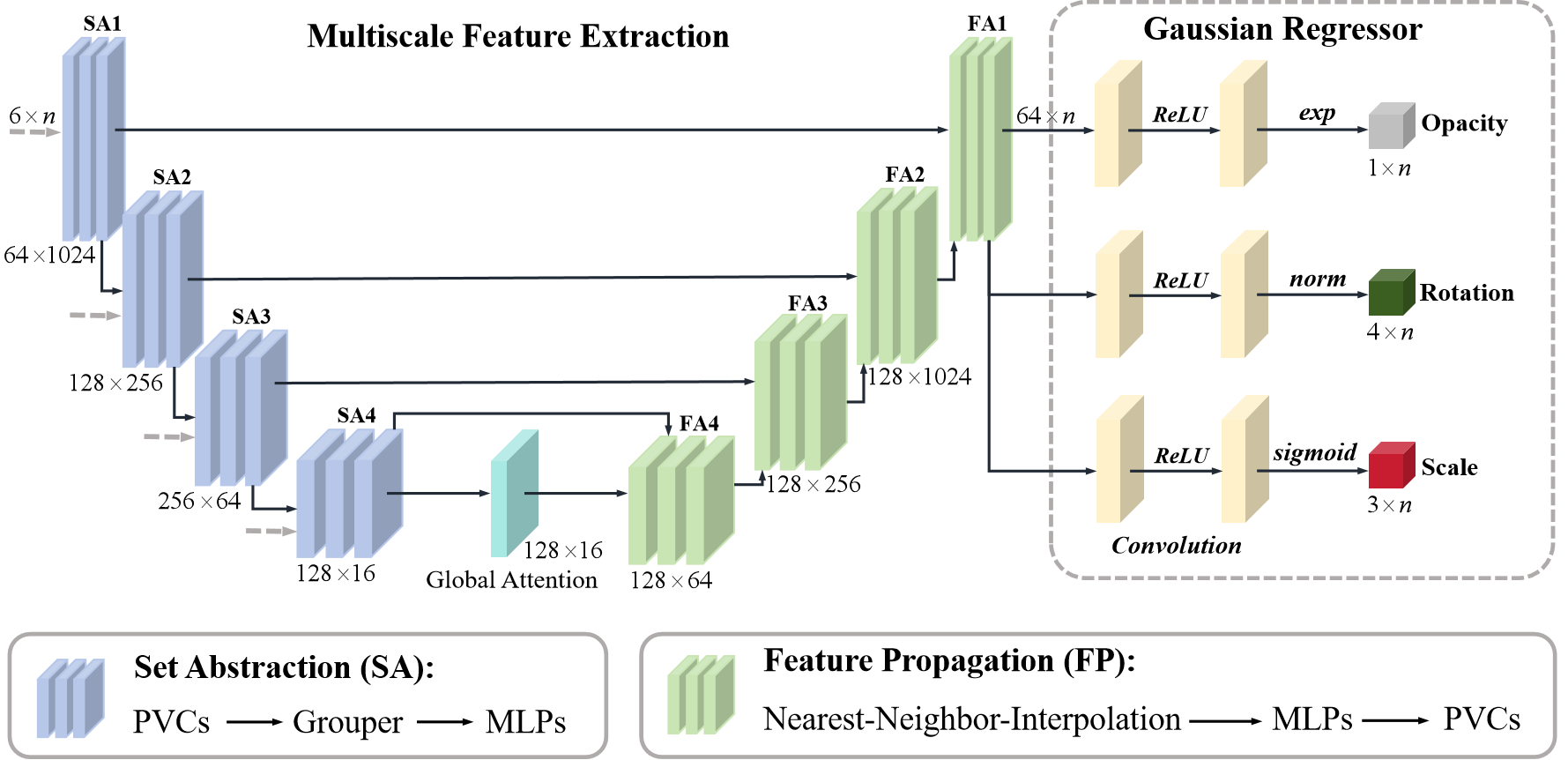}
    \caption{The architectures of multiscale feature extractor and Gaussian regressor.}
    \label{fig: pvcnn}
    \vspace{-6mm}
\end{figure}

As shown in Fig. \ref{fig: pvcnn}, the multiscale feature extraction module has four set abstractions (SA) and four feature propagations (FP) including multilayer perceptron (MLP), point-voxel convolution (PVC) \cite{liu2019point}, Grouper block (in SA) \cite{qi2017pointnet++}, and Nearest-Neighbor-Interpolation (in FP). Then, neural descriptors are extracted from color point clouds of different scales through the above modules.
% The Grouper block (in SA) consists of the sampling layer and grouping layer introduced by PointNet++.
The right side of Fig. \ref{fig: pvcnn} shows that the Gaussian regressor contains three independent heads, such as convolutions and corresponding activation functions, to predict the Gaussian properties from the obtained neural descriptors, namely rotation quaternion ($R$), scale factor ($S$), and opacity ($\alpha$), for Gaussian rasterization.

\begin{figure}
    \centering
    \includegraphics[width=\textwidth]{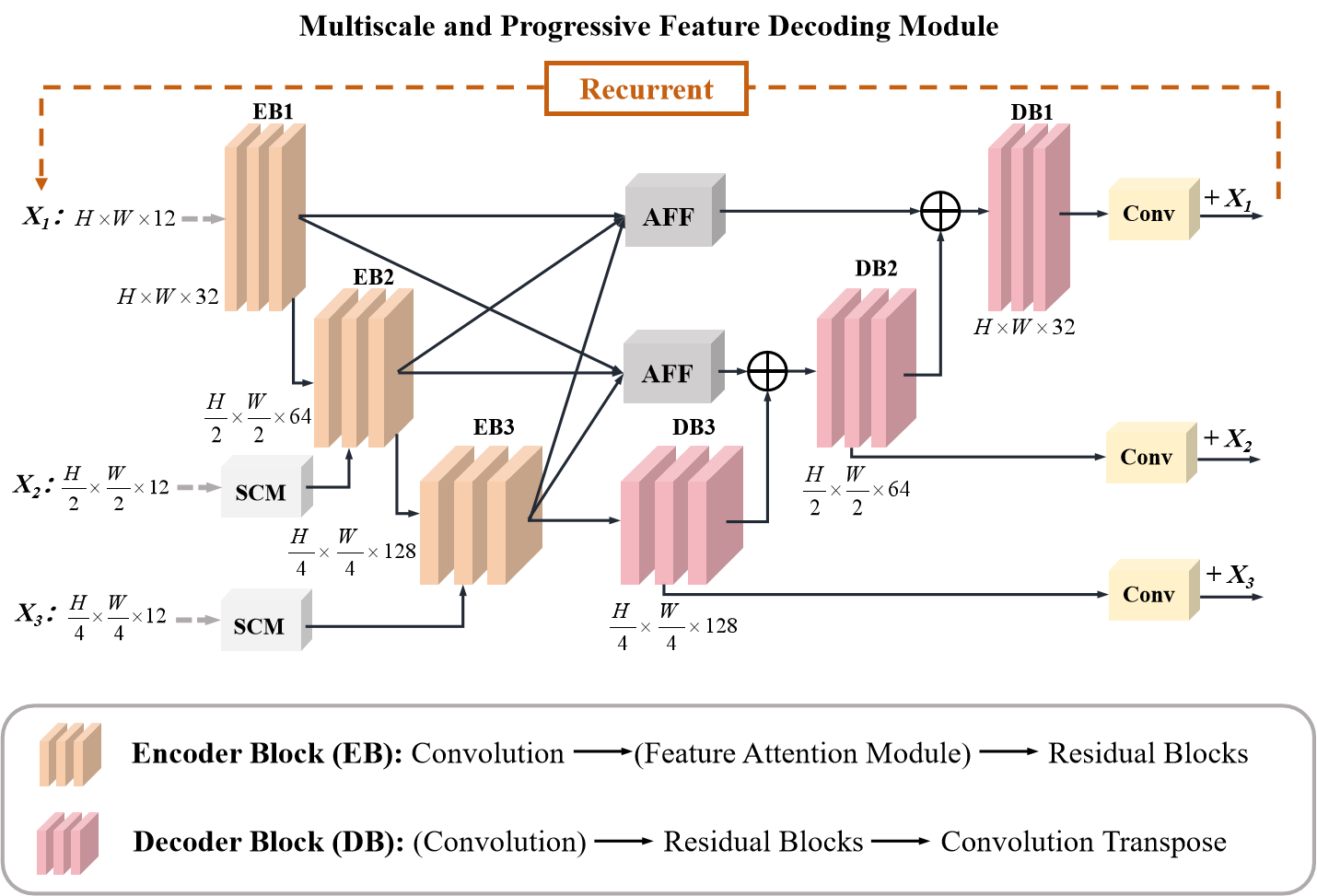}
    % \vspace{-5mm}
    \caption{The architectures of multiscale and progressive feature decoding module.}
    \label{fig: mimo}
    \vspace{-2mm}
\end{figure}
\vspace{-2mm}

\subsection{Multiscale and Progressive Feature Decoding}
Fig. \ref{fig: mimo} shows the specific network structure of the module.
$X_1$ is the input of this module and concatenates the Gaussian splatting image (3-dimensional) and the feature splatting map (9-dimensional) obtained by stage I. The multiscale and progressive feature decoding module has three encoder blocks (EB) and three decoder blocks (DB). 
In EB1 and EB2, the modules in parentheses are not used.%including components include convolution, residual blocks, feature attention module (in EB) \cite{}, and convolution transpose (in DB). 
Shallow Convolutional Module (SCM) extracts features from the downsampled input. Asymmetric Feature Fusion (AFF) means that features of multiscale are fused. Details about the above submodules introduced by MIMO-UNet \cite{cho2021rethinking}.

We recurrently run the decoder network twice. The first run of the network outputs 12-dimensional features as inputs to the second run of the network. The second output is the final rendered image (the first 3-dimensional).
%%%%%%%%%%%%%%%%%%%%%%%%%%%%%%%%%%%%%%%%%%%%%%%%%%%%
\section{More Comparative Experiments}
\subsection{More Comparisons with Other Novel View Synthesis Methods on Generalization and Per-scene Optimization}
\subsubsection{DTU.} We compare the proposed method, Trivol\cite{hu2023trivol}, and the generalized NeRF-based synthesis methods on the DTU dataset. The proposed method and Trivol are trained and tested on the original point clouds to achieve novel view synthesis in different scenes. IBRNet\cite{wangIbrnetLearningMultiview2021} and MVSNeRF \cite{chenMvsnerfFastGeneralizable2021} combine image prior with NeRF \cite{mildenhall2021nerf}. For a fair comparison, we pretrain all methods on the same training set. 

\begin{figure}[t]
 % \vspace{-4mm}
    \centering
    \includegraphics[width=\textwidth]{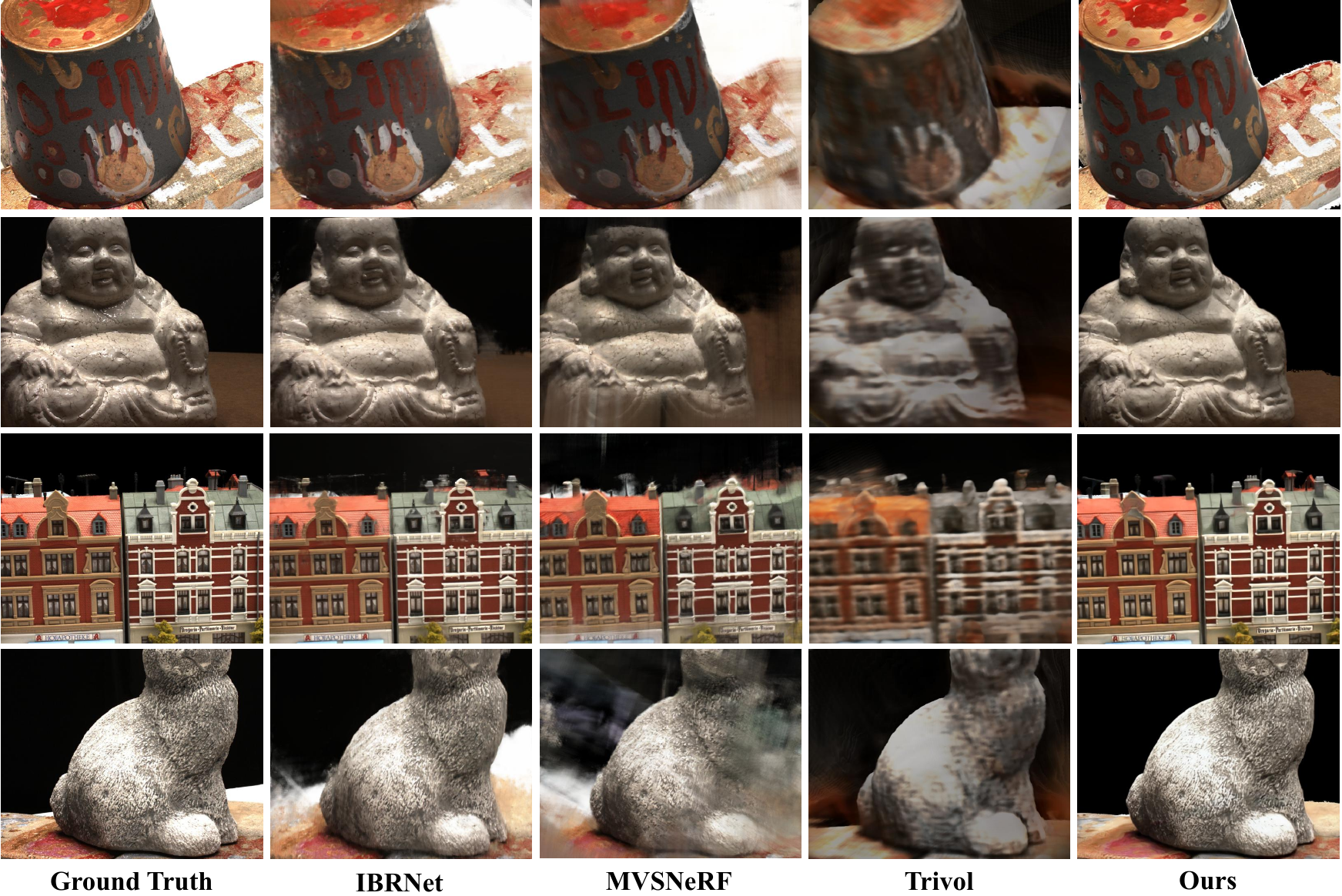}
    \caption{Qualitative comparisons between ours and other neural rendering methods on the DTU dataset.}
    \label{fig: dtu_sup}
    \vspace{-4mm}
\end{figure}

Table. \ref{Tab: dtu_sup} summarizes the quantitative comparison of PSNR, SSIM \cite{wang2004image}, and LPIPS \cite{zhang2018unreasonable}. This proves that our method is significantly superior to the joint image prior and state-of-the-art joint point cloud prior methods on all metrics.
Fig. \ref{fig: dtu_sup} shows the results of qualitative comparisons.
We can intuitively find that other methods produce severe artifacts at places with the boundaries of the object, especially the first row of red brick and the top of the house in the third row. That is because they depend on the fully implicit neural representation. However, our method is based on 3DGS and supplemented by the network design.
Our approach achieves the best visualization results both at the edges and in the local details of the object.
\begin{table}[t]
\caption{Quantitative comparisons between ours and other neural rendering methods on the DTU dataset.}
\renewcommand{\arraystretch}{1.2}
\centering
\setlength\tabcolsep{6mm}{
\begin{tabular}{lcccc}
\hline\hline
Method  & IBRNet & MVSNeRF  & Trivol  & Ours    \\ \hline
PSNR↑   & \underline{25.17} & 23.50 & 20.02  & \textbf{25.98}  \\
SSIM↑   & \underline{0.902} & 0.818 & 0.674  & \textbf{0.919}  \\
LPIPS↓  & \underline{0.181} & 0.314 & 0.483  & \textbf{0.131}    \\
\hline\hline
\end{tabular}}
\label{Tab: dtu_sup}
% \vspace{-4mm}
\end{table}

\subsubsection{ScanNet.} Table.~\ref{Tab: scannet_f_g} illustrates the quantitative comparisons between the proposed method and other SOTA methods, including two NeRF-based generalizable methods (IBRNet\cite{wangIbrnetLearningMultiview2021} and ENerF \cite{lin2022efficient}) and two per-scene optimized methods (3DGS \cite{kerbl20233d} and InstantNGP \cite{muller2022instant}). 
Due to our method of rendering photorealistic images by querying any given point clouds with arbitrary camera viewpoints in a generalizable manner, it does not require per-scene optimization. However, 3DGS and InstantNGP must be trained in a per-scene manner and cannot be generalized to unseen scenes. Therefore, We only adopt the multiview images in (scene0581\_01) ScanNet to train them and evaluate their performance on the test views. Both of the two generalizable methods receive multiview images as input and can generate novel views without per-scene optimization. 
% result performance
% For generalizable methods, 
It is clear that our method gets the best result compared to all other methods. After fine-tuning, our result is better than the result on generalization and significantly superior to two per-scene optimized methods.
For per-scene optimization

\begin{table}[t]
\caption{Quantitative comparisons between ours and other NeRF-based generalizable rendering methods and novel view synthesis methods on the ScanNet dataset.}
\renewcommand{\arraystretch}{1.2}
\centering
\setlength\tabcolsep{3mm}{
% \scalebox{0.75}{
\begin{tabular}{lcccccc}
\hline\hline
Setting    & \multicolumn{3}{c}{Generalization} &\multicolumn{3}{c}{Per-scene Optimization} \\
Method    & IBRNet& ENeRF & Ours  & InstantNGP  & 3DGS & Ours\\ \hline
PSNR↑    & 20.72 & 23.72 & \textbf{25.44} & 16.10 & 24.34 & \textbf{26.90}\\
SSIM↑    & 0.734 & 0.882 & \textbf{0.901} & 0.557 & 0.796 & \textbf{0.851} \\
LPIPS↓   & 0.515 & 0.374 & \textbf{0.164} & 0.684 & 0.419 & \textbf{0.299} \\
\hline\hline
\end{tabular}}
\vspace{-4mm}
\label{Tab: scannet_f_g}
\end{table}

\subsection{Comparisons with Point2Pix} 
We reimplement this method because the source code of Point2Pix \cite{hu2023point2pix} is not publicly available. The quantitative results of Point2Pix and Ours are shown in Table. \ref{Tab: Point2Pix}. 
It can be seen that our method still has advantages over Point2Pix in all datasets.

\begin{table}[h]
\vspace{-4mm}
    \caption{Quantitative comparisons between our method and Point2Pix on three datasets.}
    \vspace{-3mm}
    \renewcommand{\arraystretch}{1.2}
    \centering
    \setlength\tabcolsep{3mm}{
    \begin{tabular}{l|cc|cc|cc}
    \hline\hline
    Dataset &\multicolumn{2}{c|}{ScanNet} &\multicolumn{2}{c|}{DTU} & \multicolumn{2}{c}{THuman2.0}  \\ 
    Method         & Point2Pix & Ours & Point2Pix & Ours & Point2Pix & Ours \\ \hline
    PSNR↑    & 18.47& \textbf{19.86} &   21.76& \textbf{25.44}  &   31.89& \textbf{34.74}   \\
    SSIM↑    &  0.723& \textbf{0.758} &   0.871& \textbf{0.901} &   0.976& \textbf{0.983}   \\
    LPIPS↓   &   0.483& \textbf{0.452} &   0.278& \textbf{0.164}&   0.028& \textbf{0.009}  \\
    \hline\hline
    \end{tabular}}
    \label{Tab: Point2Pix}
    \vspace{-7mm}
\end{table}

\subsection{Discussions of Incomplete and Uneven Point Clouds}
The point clouds are imperfect and super sparse due to the realistic LiDAR Scanner, which often contains holes and artifacts. Our method is robust to the incompleteness and unevenness of point clouds. Here we provide some visual examples to illustrate this. 

\begin{figure}
\vspace{-4mm}
    \centering
    \includegraphics[width=1\textwidth]{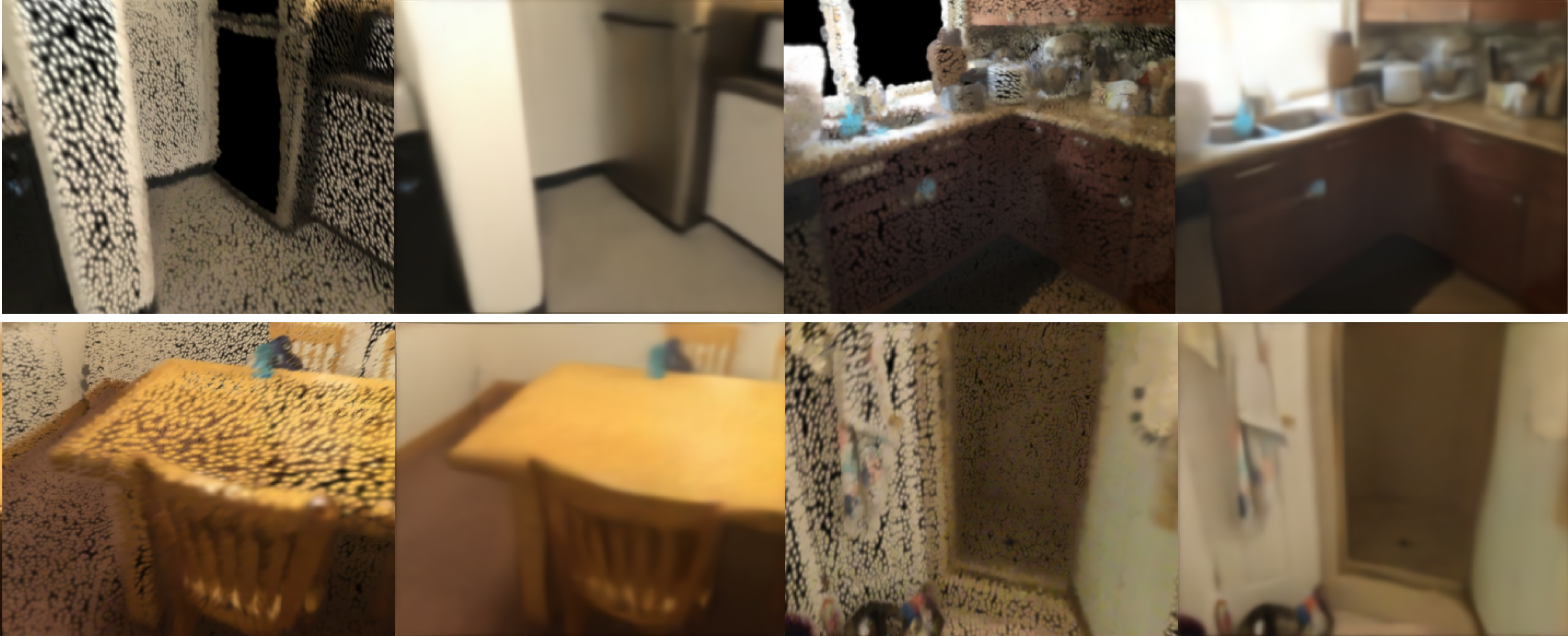}
    \caption{Examples of point clouds containing large holes (row 1) and results with uneven density (row 2). The left side of each scene is the point splatting, and the right side is our method.}
    \label{fig: holes and non}
    \vspace{-7mm}
\end{figure}

\noindent \textbf{Large holes:}. As shown in the first row of Fig. \ref{fig: holes and non},  there are huge holes in the refrigerator and the glass, and our method can well fill the huge holes.

\noindent \textbf{Non-uniform density:}
As shown in the second row of Fig. \ref{fig: holes and non}, the points are distributed unevenly within the scene. For example, points on the table are sparse, while those on the chair are more dense. It can be seen that our method still generates high-quality images in non-uniform situations. 

\begin{table}[t]
\caption{Ablation studies of the effect of different numbers of points on DTU dataset, where the metric is PSNR.}
\renewcommand{\arraystretch}{1.2}
\centering
\setlength\tabcolsep{4.2mm}{
\begin{tabular}{lcccccccc}
\hline\hline
Train   & 0.2 &  0.2  & 0.2  & 1 & 1  & 1   \\ \hline
Test    & 0.2 &  0.5  & 1 &  0.2 &  0.5  & 1  \\ \hline
NPBG++ & 18.74 & 18.27 & 17.05  & 19.40 & 21.24 & 23.14  \\
Trivol & 16.28 & 15.68 & 15.43  & 15.78 & 16.42 & 17.48  \\
Ours   & 22.34 & 22.19 & 22.05  & 12.78 & 17.74 & 25.88   \\
\hline\hline
\end{tabular}}
\label{Tab: number of points2}
\vspace{-4mm}
\end{table}

\section{More Ablation Studies}
\subsection{Further Discussions of the Generalization for the Different Number of Points}
In this section, we further analyze and evaluate how different training and test numbers of points affect performance. Even though we have already tested this point on THuman2.0 in the main text. However, the THuman2.0 is a human body dataset with a small number of points. We set more extreme training and test pairs to show the performance of our method, Trivol \cite{hu2023trivol}, and NPBG++ \cite{rakhimov2022npbg++} respectively on the DTU \cite{jensen2014large} dataset. Because DTU dataset includes more objects with different scales and different numbers of points. The number of points used for comparison in the DTU dataset is much larger than the setup in the THuman2.0 dataset.
We train the model on the downsampled point cloud by a factor of 0.2. After training, we evaluate their performance on the point cloud with different downsampling factors including 0.2, 0.5, and 1.

Table. \ref{Tab: number of points2} reports the results of this ablation, where the first two lines represent the training and testing downsampling factors. We can vividly notice that when there is a large gap in the number of points, models trained on a smaller number of points have difficulty in extracting the features and 3d contextual information of the points well when faced with a test scene with more points. For the above problems, our model is more robust to the changes in point cloud densities. However, the performance of NPBG++ and Trivol is significantly reduced. Our approach achieves the best results with the lowest performance degradation. The model on the training set with many points is also difficult to be backward compatible with the point cloud drop sampling with a large gap. However, the drawback of our model is that it cannot generalize well to fewer points if trained on dense point clouds. This might be because the Gaussian regressor learns to predict relatively small values of those Gaussian parameters when training on the high point density data. However, if the point cloud density drastically decreases, the small Gaussian parameters cannot cover the entire image plane, thereby yielding more holes and artifacts. Nevertheless, we would like to claim that our approach mainly concentrates on the scenarios of sparse point cloud density, thus this drawback could be tolerant. 

\subsection{Discussions of the Impact of Different Noises}
% As for the THuman2 dataset, the points are obtained from the depth groundtruth. THuman2 provides 3D meshes which we use to render multiview images and depths. 
The accurate points help us to explore the impact of different noises on the model robustness by adding noises to the clean point clouds. Therefore, We assess the noise's influence by adding different noise levels to the THuman2 dataset. The results of Ours Trivol are listed in Table \ref{Tab: noise}, where $\sigma$ is the variance of Gaussian noises. 
The results show that our method is robust to noises compared to Trivol. 

\begin{table}[h]
\vspace{-4mm}
\caption{Ablation studies of the impact of different noises on THuman2 dataset, where the metric is PSNR.}
% \caption{Quantitative Results }
\vspace{-1mm}
\renewcommand{\arraystretch}{1.2}
\centering
\setlength\tabcolsep{6mm}{
% \scalebox{0.8}{
\begin{tabular}{cccc}
\hline\hline
$\sigma$    & 1e-3         & 2e-3          & 3e-3    \\ \hline
Trivol    & 23.81 &21.64 & 20.04  \\
Ours    & 32.66& 31.98 & 30.85  \\
\hline\hline
\end{tabular}}
\vspace{-4mm}
\label{Tab: noise}
\end{table}

\subsection{More Visualization Comparisons with Baselines}
\begin{figure}
\vspace{-4mm}
    \centering
    \includegraphics[width=\textwidth]{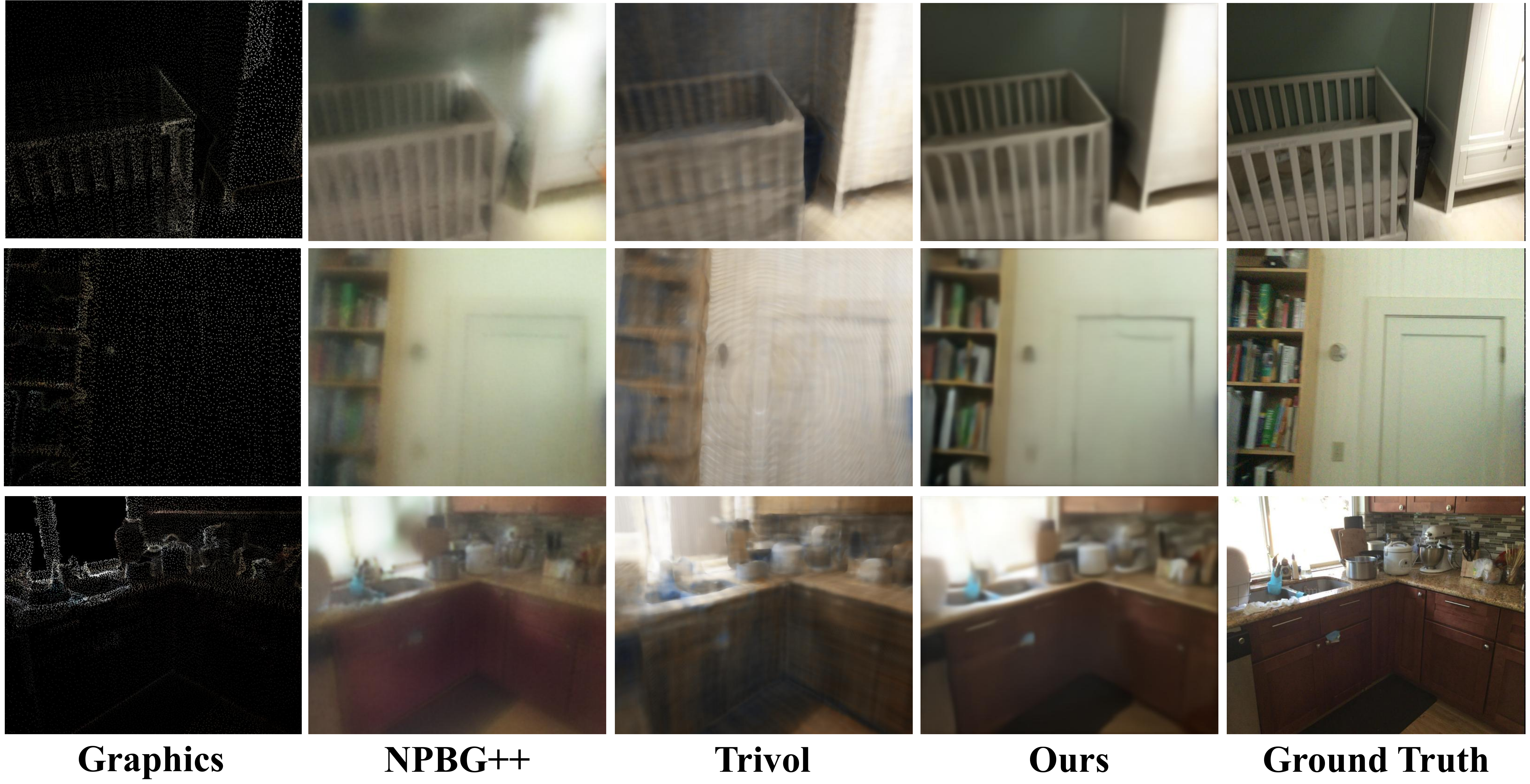}
    \vspace{-4mm}
    \caption{Additional qualitative results between ours and baselines on ScanNet dataset.}
    \label{fig: scannet_sup}
    \vspace{-10mm}
\end{figure}

\begin{figure}
\vspace{-10mm}
    \centering
    \includegraphics[width=\textwidth]{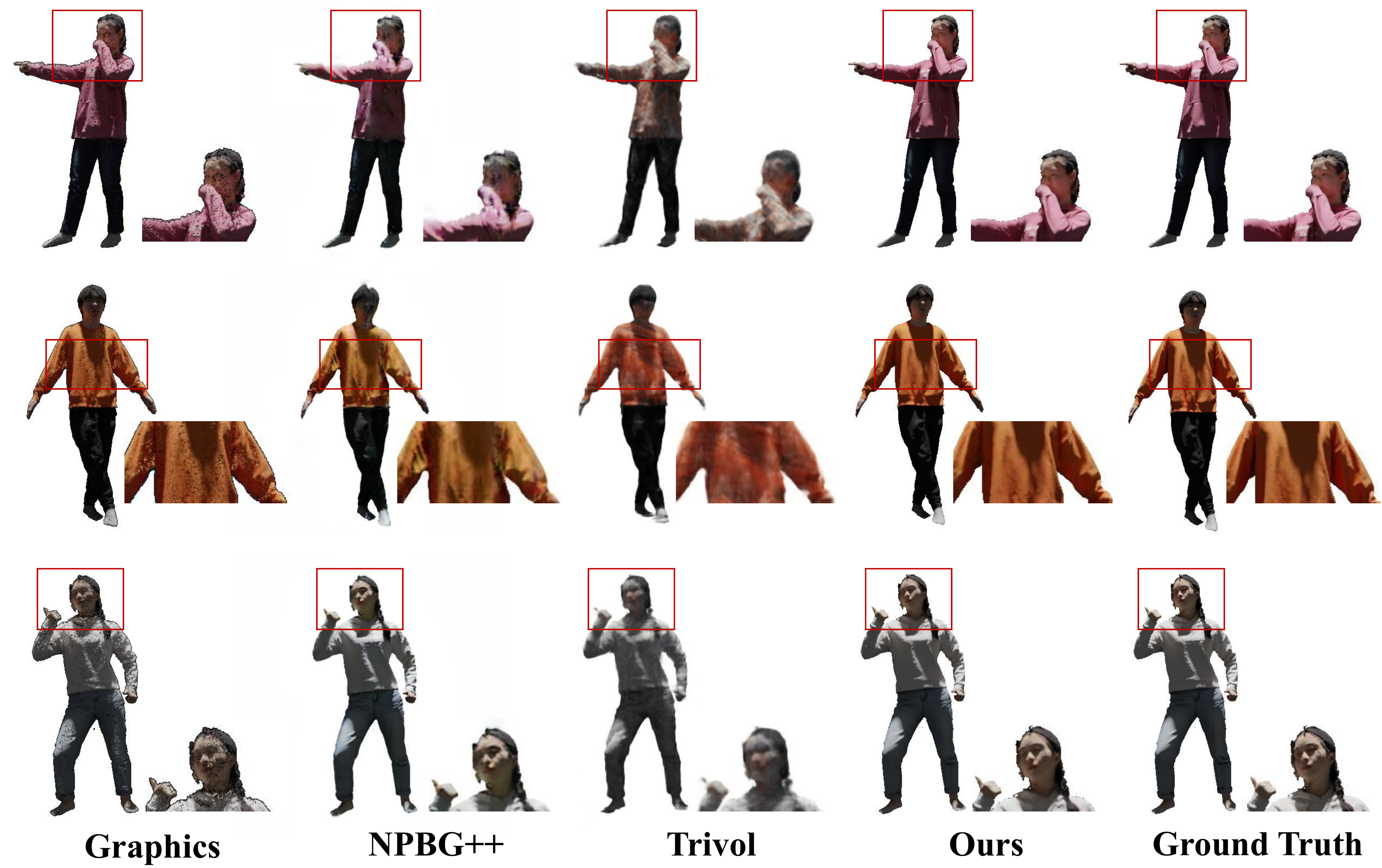}
    \caption{Additional qualitative results between ours and baselines on the THuman2.0 dataset with enlarged details.}
    \label{fig: thuman_sup}
    \vspace{-5mm}
\end{figure}

Fig. \ref{fig: thuman_sup} and Fig. \ref{fig: scannet_sup} show the additional qualitative results on ScanNet \cite{dai2017scannet} and THuman2.0 \cite{tao2021function4d} datasets with enlarged details. Our method achieves the best rendering results. Especially, at geometry discontinuity regions, our model can produce clearer and sharper boundaries than all baselines. The graphics transformation results in column 1 show the sparsities of the point cloud skeletons. These results further effectively indicate the superiority of our model when dealing with super sparse point cloud data. For example, fences and small kitchen utensils are faithfully reconstructed by our approach whereas only faint shallows appear in counterparts. Likewise, Our method delineates the expressions of the people and the wrinkles in the clothes.

\end{document}